\pdfoutput=1
\documentclass[publicdomain]{eptcs}
\providecommand{\event}{FMAS} 

\usepackage{iftex}
\usepackage{amssymb}
\usepackage{svg}
\usepackage{amsmath}
\usepackage{graphicx}
\usepackage[english]{babel}
\usepackage{dirtytalk}
\usepackage[inline]{enumitem}

\usepackage{float}

\ifpdf
  \usepackage{underscore}         
  \usepackage[T1]{fontenc}        
\else
  \usepackage{breakurl}           
\fi

\title{Verifying Safety of Behaviour Trees in Event-B}

\author{Matteo Tadiello \qquad\qquad Elena Troubitsyna
\institute{KTH Royal Institute of Technology\\ Stockholm, Sweden}
\email{tadiello@kth.se \qquad\qquad\qquad elenatro@kth.se}
}

\def\titlerunning{Verifying Safety of Behaviour Trees}
\def\authorrunning{Matteo Tadiello \& Elena Troubitsyna}

\begin{document}
\maketitle

\begin{abstract}
Behavior Trees (BT) are becoming increasingly popular in the robotics community. The BT tool is well suited for decision-making applications allowing a robot to perform complex behavior while being explainable to humans as well. Verifying that BTs used are well constructed with respect to safety and reliability requirements is essential, especially for robots operating in critical environments. In this work, we propose a formal specification of Behavior Trees and a methodology to prove invariants of already used trees, while keeping the complexity of the formalization of the tree simple for the final user. Allowing the possibility to test the particular instance of the behavior tree without the necessity to know the more abstract levels of the formalization.
\end{abstract}

\section{Introduction}

Autonomous Systems (AS) like Humanoid Robots, Autonomous Vehicles, or Unmanned Aerial Vehicles are becoming increasingly complex and need to interact with dynamic environments and with each other. For this reason, robots require tools to enable advanced perception and understanding of the environment, or capabilities to operate in complex situations. Artificial Intelligence is extending the capability of perception and action of the agents and allows robots to operate in environments not suitable for robots just a few years ago.

In most common scenarios the complexity of the environment requires to the robot to have different skills, the capability of different actions, and hence also a certain degree of reasoning and understanding of which action to take and when. A relevant example could be an urban road, with car, pedestrian, and signals. That's why planning becomes more important with the increasing complexity of the Operational Design Domain (ODD). Various techniques have been developed to solve this challenge. They include  a variety of machine learning approaches as well as 
Behaviour Trees. Behaviour Trees started as a tool for video-game development of Non-Playable Character \cite{10.1145/2987491.2987513}, and are now being applied also to robotics due to their flexibility and high usability.

Usually, robotics systems should be dependable, i.e.,   being able to operate without putting at risk others, or themselves, while continuing to provide their service in a reliable way. Formal approaches can be used to provide these characteristics and ensure the safety and reliability of the systems. Applying formal methods to AI techniques, however, is still challenging given the enormous search space and states possibility, and became particularly relevant when dealing with Neural Networks with billions of parameters.
The importance of safety in robots that need to operate in a day-to-day basis in a critical environment is extremely important, as well as the capability of the system to explain why certain decisions have been made. For this reason Behaviour Trees (BT) have become particularly interesting in the robotics industry. This mathematical model follows simple principles and allow readability, modularity and re-usability. And for this reason, it has gain popularity especially in decision making tasks. 

However, even with these properties BT can become quite complex and easy to contain design mistakes that could mine the safety and reliability of the system. For these reasons, it is important to ensure that certain properties are kept during the execution of the BT that ensure the dependability of the system built. The usage of formal methods to ensure the correctness of BT is still scarce, and difficult to build. 

In this work, we try to advance in this direction by providing a formal specification of the Behaviour Tree framework that will allow the verification of correctness of specific BT instances and holding of interesting invariant properties that the system should maintain. In particular, we are able to define safety invariants and prove that these hold for specific BTs, while maintaining the re-usability and modularity of the tree structure.
The system modeling has been built using the Event-B modeling method~\cite{abrial_2010} and the Rodin~\cite{RODINPLAT} platform as a support for the refinements and mathematical proofs.

The rest of this paper is structured in the following way: we first provide an introductory explanation of Behaviour Trees and Event-B in the Background section, we continue with the Method section presenting the BT formal specification, followed by a case study section where we show a simple BT instance and the method to convert a BT to a formal specification that can be checked. We will then move to the related work and end with the conclusion section.

\section{Background}

\subsection{Behaviour Trees}
At a high level, a Behaviour Tree is a mathematical model that structures the switching of different tasks, which a virtual agent or a robot can perform~\cite{colledanchise_behavior_2018}.
This model uses a rooted tree where every node has a single parent and no loops.
The nodes in a BT can be subdivided into three categories:
\begin{itemize}
    \item \textbf{root}: the root node. It does not have parents, and it normally has just one child.
    \item \textbf{internal node}: also known as control flow nodes, those are nodes that are at the intermediary level of the tree. They all have a parent and at least a child. They represent how to navigate the tree.
    \item \textbf{leaf node}: also known as execution nodes, these nodes have a parent but do not have a child. They represent the tasks that the agent needs to perform (like "go to point A") or conditions (like "battery > than 30\%) that needs to be true before executing an action.
\end{itemize}

\begin{figure}[t!]
    \centering
    \vspace{0.4cm}
    \begin{center}
    \includegraphics[width=14cm]{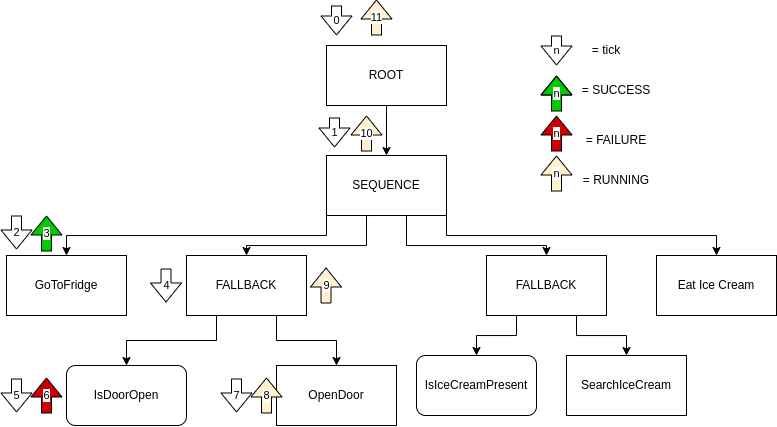}
    \caption{A sample of a BT and its ticking process. The execution of the tree returns RUNNING while executing the action of opening the door.}
    \label{fig:BT_example}
    \end{center}
\end{figure}

In practice, the BT is navigated from the root to the leaves to decide which task needs to be executed at every time. 
To navigate the tree a process called \emph{ticking} is used. An example of a BT and its ticking process can be seen in Fig~\ref{fig:BT_example}. The navigation starts by ticking the root and then recursively ticking the left-most child of the node. When a leaf node is ticked it immediately replies with the status and sends the signal back to its parents, then the continuation of the navigation is pursued by following the rules of the internal nodes. A node is executed only if it has been ticked.
At each time step, the tree is explored and returns the execution status of the agent. The output value can be $SUCCESS$, $RUNNING$, or $FAILURE$ which, respectively, represent if the behavior of the agent, is in a success state, in a failure, or if it is still running. We call this the status of the tree. 

The status of the tree is due to the composition of the statuses of the nodes in each subtree analyzed, and each node follows different rules to provide its status.

More specifically, for the leaf nodes we can distinguish two kinds:
\begin{itemize}
    \item \textbf{Conditions}: these nodes represent guards check, they are normally placed before tasks nodes, they can provide just $SUCCESS$ or $FAILURE$, but not $RUNNING$)
    \item \textbf{Tasks} or \textbf{Actions}: these nodes represent the actual actions executed by the agent. Once ticked, they immediately return SUCCESS, RUNNING, or FAILURE.
\end{itemize}

The leaf nodes are the ones that check the status of the agent and execute actions. The internal nodes instead define how leaf nodes are, or are not, executed.

The internal nodes, analyzed, are divided into two categories:
\begin{itemize}
    \item \textbf{Sequence}: these nodes execute a series of children in sequence. The next child is executed only if the previous child has returned SUCCESS. The sequence node return to their parent SUCCESS if and only if all the children have returned SUCCESS. It returns RUNNING if one of the children is RUNNING, and FAILURE otherwise (hence one of the children has returned FAILURE). The sequence symbol is a box with the "$\rightarrow$" label.
    \item \textbf{Fallback}: these nodes tick a series of children following a left-to-right order, starting from the leftmost child. The next child is executed only if the previous child has returned FAILURE. The fallback node return to their parent FAILURE if and only if all the children have returned FAILURE. It returns RUNNING if one of the children is RUNNING, and SUCCESS otherwise (hence one of the children has returned SUCCESS). 
    The fallback symbol is a box with the "$?$" label.
\end{itemize}

BT can be extended also with other two internal nodes categories called PARALLEL and DECORATORS. Those nodes, less frequent, are not treated in this work.

The main strengths of Behaviour Tree are that they are simple and readable, and each subtree is reusable given the modularity property of this structure. But they can represent also complex behaviors for robots operating in challenging environments. 
Moreover, BT can also be extended with different properties like Stochastic BTs~\cite{stocastic_BT} or automatically generated as in~\cite{DBLP:journals/corr/ColledanchisePO15}.

\subsection{Event-B and Rodin}
\begin{figure}[t!]
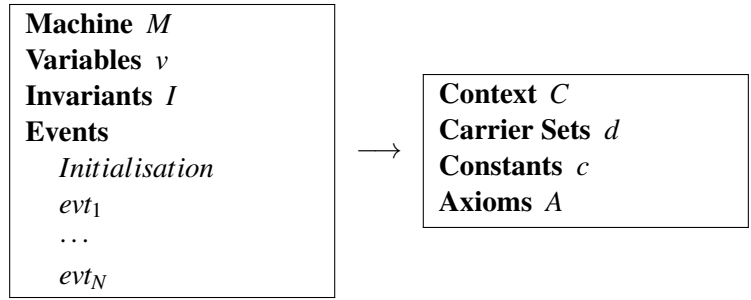

    \centering
\vspace{0.4cm}
\begin{center}
\fbox{
\parbox[center]{4cm}{
\textbf{Machine} $\;M$ \\
\hspace*{0pt}\textbf{Variables} $\; v$ \\
\hspace*{0pt}\textbf{Invariants} $\; I$ \\
\hspace*{0pt}\textbf{Events} \\
\hspace*{10pt} $Initialisation$ \\
\hspace*{10pt} $evt_1$ \\
\hspace*{10pt} $\cdots$ \\
\hspace*{10pt} $evt_N$
}}
\hskip 5pt
$\longrightarrow$
\hskip 5pt
\fbox{
\parbox[center]{4cm}{
\textbf{Context} $\;C$ \\
\hspace*{0pt}\textbf{Carrier Sets} $\; d$ \\
\hspace*{0pt}\textbf{Constants} $\; c$ \\
\hspace*{0pt}\textbf{Axioms} $\; A$
\smallskip
}}

\caption{Event-B machine and context}
\label{fig:am}
\end{center}
\vspace{0.2cm}
\end{figure}

The formal development framework used in this work is the \emph{Event-B} formalism. In Event-B a system specification is defined using the concept of \emph{abstract-state machine}. This model encapsulates a collection of variables and contains the set of operations to interact with the states of the machine called \textit{Events}. We can thus define the behavior of the machine by describing the dynamic part of the state machine. 
Another component of the Event-B method is the so-called \textit{context}, these components contain the static part of the system. For instance, in the context user-defined types are defined or constant components of the system. The context can be seen by the abstract machines as shown in \ref{fig:am}.

The dynamic part of the system is represented by the variables of the abstract-state machines and the Cartesian product of their range represents the state space of the machine. The variables are strongly typed by invariant statements and are initialized using a special initialization event. The invariant statements, not only define the type of the variables but most importantly, are used to define properties that the system should hold for the entire time of execution. For instance, a possible invariant could be a safety property like: "the distance to the obstacles needs to be at least 5 meters from the robot".

The changing of the variables, and hence the state of the machine, is represented through a set of Event clauses defined in the machine. 
An event is normally defined as follows:

\begin{displaymath}
    \mathbf{evt}\; \widehat{=}\; \mathbf{any }\; variables\; \mathbf{where}\; guards\; \mathbf{then}\; S\; \mathbf{end}
\end{displaymath}
where variables is a new list of local variables, guards are predicates defining when the event can be triggered, and $S$ is a set of statements or assignment that changes the state of the machine. However, both the variables and the guards are optional in the definition of an event.
Multiple events can be enabled at the same time, in this case, any of them can be chosen to be executed in a non-deterministic way.

One of the strong points of Event-B is that it allows the possibility of machine refinement. This allows us to start the definition from a very high and abstract level, and include more details, variables, and events at every refinement. We can then introduce implementation details in steps while preserving functional correctness.  

The semantics of an Event-B model is formulated as a collection of proof obligations \cite{abrial_2010}, and the process of verification and generation and proving of proof obligations, has been facilitated by the Rodin platform \cite{RODINPLAT}.

\section{Specification of Behaviour Trees in Event-B}

The formal specification of a generic Behaviour Tree has been implemented and proven using Event-B and Rodin\footnote{The source code can be found here: \url{https://github.com/tadteo/BT_eventb}}.
To do that, we have used 4 different refinements. We built initially the abstract figure of the nodes, till arriving at the instances of the BT, which can be found as a refinement of the 4th level.
Each abstract machine is coupled with a corresponding context.
The four abstraction levels are described in detail in the following subsections, and are:
\begin{itemize}
    \item Nodes
    \item Tree
    \item Behavior Tree
    \item Specific Instance of a BT.
\end{itemize}

 \subsection{Specifying a node}
 We start defining the nodes and certain properties that the nodes composing the Tree need to have. At this level, we are not interested in the connection and relationship between the nodes but just in their properties.
 
 First of all, we define two new types to define the kind of node and the result that each node can return.
 
 The $TYPE$ type represents the category that each node can be and it is defined as follow:
 \begin{displaymath}
 TYPE = \{ROOT, SEQUENCE, FALLBACK, CONDITION, ACTION\}
 \end{displaymath}
 It contains the types for the root, control flow nodes, and execution nodes, this will be used in a future refinement to define the different navigation behavior of the nodes.
 
 The RESULT type, instead, represents the value of return that each node can contain. We define it as follows:
 \begin{displaymath}
 RESULT = \{SUCCESS, RUNNING, FAILURE, UNKNOWN\}
 \end{displaymath}
 The first three elements are the normal return value of a node in a BT, the $UNKNOWN$ element instead is used as the default value when a node has not been ticked yet, and hence the return value is unknown.
    
We wanted to define the single nodes as a data structure containing the following members:

\begin{center}
    \begin{verbatim}
    node {
      int n_id; //constant value in context
      TYPE n_type; //constant value in context
      bool n_tick; //dynamic value in machine   
      RESULT n_result; //dynamic value in machine
    };
    \end{verbatim}
\end{center}

Where \texttt{n_id} represents a unique identifier of the node that starts from 0 with the root. This value is used to define an order on the nodes used in a future refinement to identify the order of the children, used to navigate the tree from left to right. And \texttt{n_tick} is a boolean variable to see if the node has been \textit{ticked} or not. 

Since in Event-B there is not a standard definition of a structure, the members of the nodes have been defined as relationships between the nodes and the corresponding type as follows:

\begin{center}
\fbox{
\parbox[center]{6.5cm}{
    \hspace*{0pt} $\textnormal{\texttt{n_type}} \in NODES \rightarrow TYPE$ \\
    \hspace*{0pt} $\textnormal{\texttt{n_id}} \in NODES \rightarrow \mathbb{N}$ \\
    \hspace*{0pt} $\textnormal{\texttt{n_tick}} \in NODES \rightarrow BOOL$ \\
    \hspace*{0pt} $\textnormal{\texttt{n_result}} \in NODES \rightarrow RESULT$
    }
}
\end{center}
and initialized as: 

\begin{center}
\fbox{
\parbox[center]{6.5cm}{
\hspace*{0pt} $\textnormal{\texttt{n_tick}} \in NODES \times \{FALSE\}$ \\
\hspace*{0pt} $\textnormal{\texttt{n_result}} \in NODES \times \{UNKNOWN\}$
}}
\vspace{2pt}
\end{center}
where $NODES$ is the set of all possible nodes that can be used to build the tree, and that has been defined in the node context together with the types, and the constant values. \texttt{n_id} and \texttt{n_type} does not require initialization since they are defined in the context and remain constant.

\subsection{Formal model of tree}
In the first refinement, we give the nodes a structure and build the rules to construct a tree. 
We used the following four requirements to define the tree structure:
\begin{itemize}
    \item \textbf{Req1}: There is a single root object
    \item \textbf{Req2}: Each object other than the root has a parent
    \item \textbf{Req3}: There are no loops in the parent structure
    \item \textbf{Req4}: Each object is reachable from the root
\end{itemize}

First of all, we define the \textit{nodes} constant which represents the set of nodes (a subset of NODES) in the tree, with \textit{root} as the ROOT node. 

For \textbf{Req1} we define an axiom that states that for every pair of nodes, if one is a ROOT then the other is not ROOT:
\begin{equation}
\forall n1,n2 \cdot (n1 \in nodes \land n2 \in nodes \land n1 \neq n2 \land n\_type(n1) = ROOT \implies n\_type(n2) \neq ROOT)
\end{equation}

For \textbf{Req2} we define the relation \textit{parent} as following:
\begin{equation}
parent \in nodes\backslash \{root\} \rightarrow nodes
\end{equation}

For the no loop property we use the inverse of the parent function (i.e. the children)\footnote{In Event-B in Rodin, the inverse is written using the $\sim$ symbol.In this case, the parent function would be written as: $parent \sim [\{node\}]$} in the following way:
\begin{equation}
    \forall n \cdot (n \subseteq parent^{-1}[n] \implies n = \emptyset)
\end{equation}

For \textbf{Req4} we use a transitive closure on a node relation, that in a graph represents a relation to any other node in the graph \cite{MCCOLL198667}.

To achieve that, we first define a generic relation between two nodes, and then we define the transitive closure of it.
\begin{center}
\fbox{
\parbox[center]{13cm}{
\hspace*{0pt} $node\_rel = NODES \leftrightarrow NODES$ \\
\hspace*{0pt} $tcl \in node\_rel \rightarrow node\_rel$ \\
\hspace*{0pt} $\forall r \cdot ( r \in node\_rel \implies r \subseteq tcl(r)$ \\
\hspace*{0pt} $\forall r \cdot ( r \in node\_rel \implies r;tcl(r) \subseteq tcl(r)$ // The unfolding of tcl is also part of tcl\\
\hspace*{0pt} $\forall r,t \cdot ( r \in node\_rel \land r \subseteq t \land r;t \subseteq t  \implies tcl(r) \subseteq t$ //tcl(r) is least \\
\hspace*{0pt} $\forall r \cdot ( r \in node\_rel \implies r \cup (r;tcl(r))$
}}
\end{center}

\subsection{Behavior Tree level}
The Behavior Tree level starts from the definition of the tree present in the level above and adds all the rules to allow the correct navigation and return of the result.

In this level, we simulate the ticking process, which starts from a completely unvisited and unticked tree. As the first step, the root needs to be ticked. After this, a chain reaction of events brings to the exploration of the tree which eventually returns a result. The exploration ends whenever the result arrives at the root node. At this point, the tree got reinitialized and the root get ticked again, restarting the cycle.

To achieve correct navigation of the tree, each type of node needs to be treated separately with specific events.

\subsubsection{Root related events}
To deal with the root we have 3 distinct events:  

The first \textbf{tick_root} is used to tick the root node. It is used whenever the tick status of the root is FALSE and the result is $UNKNOWN$:

\begin{center}
   \begin{small}
        \fbox{
        	\parbox[center]{15.2cm}{
            \vspace{1pt}
        	\hspace*{0pt} $\mathsf{tick\_root} \;\widehat{=}$ \\
        	\vspace{1pt}
        	\hspace*{5pt} $\textbf{any} \; \; node$\\
        	\hspace*{5pt} $\textbf{where} \; \;$\\
            \hspace*{10pt} $n\_tick(node) = FALSE $\\
            \hspace*{10pt} $n\_type(node) = ROOT$ \\
        	\hspace*{10pt} $n\_result(node) = UNKNOWN$ \\
        	\hspace*{10pt} $node \in NODES$ \\ 
        	\hspace*{10pt} $node \in nodes$ \\
        	\hspace*{5pt} $\textbf{then}$ \\
        	\hspace*{10pt} $n\_tick(node) := TRUE$
         }
        }
   \end{small}
\end{center}

We then have an event that is used to tick the first unticked child of the root as soon as the root is ticked. In our specification, the root has just one child and is treated as a special node. This allows a simpler way to manage the start and end of an exploration of the tree.

\begin{figure}[H]
    \centering
    \begin{center}
    \begin{small}
    \fbox{
    	\parbox[center]{15.2cm}{
        \vspace{1pt}
    	\hspace*{0pt} $\mathsf{root\_ticked} \;\widehat{=}$ \vspace{1pt} \\
    	\hspace*{5pt} $\textbf{any}\; \;$\\
    	\hspace*{10pt} $\; \; node$\\
    	\hspace*{10pt} $\; \; child$\\
    	\hspace*{5pt} $\textbf{where} \; \;$\\
        \hspace*{10pt} $n\_tick(node) = TRUE $\\
        \hspace*{10pt} $n\_type(node) = ROOT$ \\
    	\hspace*{10pt} $n\_tick(child) = FALSE $\\
    	\hspace*{10pt} $node \in nodes$ \\
    	\hspace*{10pt} $child \in \{x | x \in parent^{-1}[\{node\}] \land n\_tick(x) = FALSE \land \\
    	\forall y \cdot( y \in parent^{-1} [\{node\}] \land n\_tick(y) = FALSE \implies n\_id(y) \geq n\_id(x) \}$ \\
    	\hspace*{5pt} $\textbf{then}$\\
    	\hspace*{10pt} $n\_tick(child) := TRUE$ \\
    	\hspace*{10pt} $analyzing\_subtree(child) := TRUE$
    	}
     }
    	
    \end{small}
    \end{center}
\caption{The event $\mathsf{root\_ticked}$}
\end{figure}

To identify the first unticked child we make use of the id of the node. The id of the nodes has been given in a way that respects the order of visiting of the tree, following a left-to-right order, with siblings having the leftmost with a smaller id and the rightmost with the highest id. When choosing which node to visit, we select as a child the node that has the \textit{n\_id} smaller than all the other nodes, and that is not yet ticked.
Finally, when the ticking process return to the root an event called \textbf{result_arrived} is used to copy the result of the child of the root to the root.

\begin{figure}[H]
    \centering
    \begin{center}
    \begin{small}
    \fbox{
    	\parbox[center]{11.8cm}{
        \vspace{1pt}
    	\hspace*{0pt} $\mathsf{result\_arrived} \;\widehat{=}$ \vspace{1pt} \\
    	\hspace*{5pt} $\textbf{any}\; \;$\\
        	\hspace*{10pt} $\; \; node$\\
                \hspace*{10pt} $\; \; child$\\
    	\hspace*{5pt} $\textbf{where} \; \;$\\
                \hspace*{10pt} $node \in NODES$ \\
                \hspace*{10pt} $n\_tick(node) = TRUE $\\
                \hspace*{10pt} $n\_type(node) = ROOT$ \\
                \hspace*{10pt} $n\_result(node) = UNKNOWN $ \\
                \hspace*{10pt} $child \in parent^{-1}[\{node\}]$ \\
                \hspace*{10pt} $child \in \{x | x \in parent^{-1}[\{node\}] \land n\_result(x) \neq UNKNOWN\}$ \\
            \hspace*{5pt} $\textbf{then}$\\
    	       \hspace*{10pt} $n\_result(node) := n\_result(child) $
    	}
    }
    \end{small}
    \end{center}
\end{figure}

 Before restarting the ticking process an event called \textbf{root_reinitialize} is used to reinitialize the tree nodes as unexplored and with unknown results.

\begin{figure}[t!]
    \centering
    \begin{center}
        \begin{small}
        \fbox{
        	\parbox[center]{15.2cm}{
            \vspace{1pt}
        	\hspace*{0pt} $\mathsf{root\_reinitialize} \;\widehat{=}$ \vspace{1pt} \\
        	\hspace*{5pt} $\textbf{any}\; \;$\\
        	\hspace*{10pt} $\; \; node$\\
        	\hspace*{5pt} $\textbf{where} \; \;$\\
            \hspace*{10pt} $n\_tick(node) = TRUE $\\
            \hspace*{10pt} $n\_type(node) = ROOT$ \\
        	\hspace*{10pt} $n\_result(node) \neq UNKNOWN $\\
        	\hspace*{10pt} $node \in NODES$ \\
        	\hspace*{10pt} $node \in nodes$ \\
        	\hspace*{5pt} $\textbf{then}$\\
        	\hspace*{10pt} $n\_result := NODES \times \{UNKNOWN\}$ \\
        	\hspace*{10pt} $n\_tick := NODES \times \{FALSE\}$
        	}
         }	
        \end{small}
    \end{center}
    \caption{The event $\mathsf{root\_reinitialize}$}
\end{figure}

\subsubsection{Leaf nodes related events}

The leaf nodes can be of two types: $CONDITION$ or $ACTION$. The leaf nodes are the only ones that directly provide a result with the ticking of the nodes. However, these events need to be refined at the next level for each case of result ($SUCCESS$, $FAILURE$ for the $CONDITION$ nodes and $SUCCESS$, $FAILURE$ and $RUNNING$ for the $ACTION$ nodes). And providing the correct logic to return the result wanted, since for each BT the logic to return the result can be different.

At this level of abstraction, the only action executed by these nodes is setting the \textit{analyzing\_subtree} variable of the parent node to \texttt{FALSE}. This is done as follows:
\begin{equation}
    analyzing\_subtree(parent(node)) := FALSE
\end{equation}

\subsubsection{Internal nodes related events}
The internal nodes are the ones in charge of controlling the flow of execution and analysis of the tree. In this work, we have analyzed the \textbf{sequence} node and the \textbf{fallback} node.
To formalize the behavior of this kind of node, we have used 5 different events for each kind covering all the possible ways in which the node can be found while analyzing the tree. At each step of the tree exploration, the control flow nodes either have enough information to return a value to the parents or need to continue the analysis of the children and tick the next child on the list.
The five events have the following form: 
\begin{enumerate*}
  \item \textit{*_ticked_initial}, 
  \item \textit{*_ticked_success}, 
  \item \textit{*_ticked_failure}, 
  \item \textit{*_ticked_running}, 
  \item \textit{*_ticked_continue}, 
\end{enumerate*}
with * used as a wildcard that can be substituted with the name of the node (like \textit{fallback_ticked_continue}).

The \textit{*_ticked_initial} event is used to start the ticking process of the children and it's only purpose is to tick the first children of these nodes.

Let's analyze now the events of the fallback node.\\
The definition is \textit{"The Fallback node [...] corresponds to routing the ticks to its children from the left until it finds a child that returns either SUCCESS or RUNNING, then it returns SUCCESS or RUNNING accordingly to its own parent. It returns FAILURE if and only if all its children return FAILURE. Note that when a child returns RUNNING or SUCCESS, the Fallback node does not route the ticks to the next child (if any)"} \cite{colledanchise_behavior_2018}. \\
To be able to route the ticks to the correct child (from left to right), we first need to correctly identify the first unticked child and last analyzed child. To do that we use a similar approach to the one used in the \textit{root_ticked} event. We use the inverse of \textit{parent} to identify the children of a node and then we use the \textit{n_id} of the children to analyze the order, finally, we use the \textit{n_result} and \textit{n_tick} variables to check the correct value we are looking for.

For instance, to identify if the last child analyzed has returned the $RUNNING$ result (used in \textit{fallback_ticked_running}, Figure \ref{fig:fallback_events}, we use the following guard:
\begin{multline}
    \exists x \cdot (x \in  parent^{-1}[\{node\}] \land n\_tick(x) = TRUE \land \\
            \forall y \cdot (y \in parent^{-1}[\{node\}] \land n\_tick(y) = TRUE \implies n\_id(y) \leq n\_id(x)) \\
            \land n\_result(x) = RUNNING)
\end{multline}
In this code, we check if exists a node of the children set, which has been ticked, that has the greatest id of all the other ticked children, and that has the result set to RUNNING.
To find the first unticked children instead we use the same approach but instead check that the \textit{n_tick} variable is set to false and that the \textit{n_id} value is the minimum of the unticked children. This can be seen in the events \textit{fallback_ticked_continue} Fig. \ref{fig:fallback_events}, and \textit{sequence_ticked_continue} Fig. \ref{fig:sequence_events},

In some events, we want also to be able to identify whenever all the children have been analyzed (ticked). In these cases, we just check that the set of children of a node that has \textit{n_tick} set to FALSE is an empty set. The guard used to check this is:
\begin{equation}
    \{x | \; x \in  parent^{-1}[\{node\}] \; \land \; n\_tick(x) = FALSE \} = \emptyset
\end{equation}

With these guards, we are now able to define the 4 events that allow us to correctly manage the Fallback node (Figure \ref{fig:fallback_events}). 

\begin{figure}[t!]
    \centering
    \begin{center}
        \begin{small}
        \fbox{
            \parbox[center]{15.7cm}{
                \hspace*{0pt} $\textbf{Events}\vspace*{7pt}$   \\
                \parbox[center]{8.27cm}{
                    \hspace*{0pt}  $ \mathsf{fallback\_ticked\_running}  \;\widehat{=}$ \vspace{1pt} \\
                    \hspace*{5pt}  $ \textbf{any} \; \; node$\\
                    \hspace*{5pt}  $ \textbf{where} $\\
                    \hspace*{10pt} $ n\_type(node) = FALLBACK $\\
                    \hspace*{10pt} $ n\_tick(node) = TRUE $\\
                    \hspace*{10pt} $ n\_result(node) = UNKNOWN $\\
                    \hspace*{10pt} $ \exists x \cdot (x \in  parent^{-1}[\{node\}] \land n\_tick(x) = TRUE \land \\
                                \forall y \cdot (y \in parent^{-1}[\{node\}] \land n\_tick(y) = TRUE \implies \\
                                n\_id(y) \leq n\_id(x)) \land n\_result(x) = RUNNING)$\\
                    \hspace*{5pt}  $ \textbf{then} $\\
                    \hspace*{10pt} $ n\_result(node) := RUNNING $ \\
                    \hspace*{10pt} $ analyzing\_subtree(parent(node)) := FALSE $ \\
                    \hspace*{5pt}  $ \textbf{end}\vspace{7pt} $ \\
                    \hspace*{0pt}  $ \mathsf{fallback\_ticked\_failure}  \;\widehat{=}$ \vspace{1pt} \\
                    \hspace*{5pt} $\textbf{any} \; \; node$\\
                    \hspace*{5pt} $\textbf{where} $\\
                    \hspace*{10pt} $n\_type(node) = FALLBACK $\\
                    \hspace*{10pt} $n\_tick(node) = TRUE $\\
                    \hspace*{10pt} $n\_result(node) = UNKNOWN $\\
                    \hspace*{10pt} $\exists x \cdot (x \in  parent^{-1}[\{node\}] \land n\_tick(x) = TRUE \land \\
                                \forall y \cdot (y \in parent^{-1}[\{node\}] \land n\_tick(y) = TRUE \implies \\
                                n\_id(y) \leq n\_id(x)) \land n\_result(x) = FAILURE)$\\
                    \hspace*{10pt} $\{x | \; x \in  parent^{-1}[\{node\}] \; \land \; n\_tick(x) = FALSE \} = \emptyset$ \\
                    \hspace*{5pt} $\textbf{then}$\\
                    \hspace*{10pt} $n\_result(node) := FAILURE$ \\
                    \hspace*{10pt} $analyzing\_subtree(parent(node)) := FALSE$ \\
                    \hspace*{5pt} $\textbf{end}$\\
                }
                \parbox[center]{8cm}{
                    \hspace*{0pt}  $ \mathsf{fallback\_ticked\_success}  \;\widehat{=}$ \vspace{1pt} \\
                    \hspace*{5pt}  $ \textbf{any} \; \; node$\\
                    \hspace*{5pt}  $ \textbf{where} $\\
                    \hspace*{10pt} $ n\_type(node) = FALLBACK $\\
                    \hspace*{10pt} $ n\_tick(node) = TRUE $\\
                    \hspace*{10pt} $ n\_result(node) = UNKNOWN $\\
                    \hspace*{10pt} $ \exists x \cdot (x \in  parent^{-1}[\{node\}] \land n\_tick(x) = TRUE \land \\
                                \forall y \cdot (y \in parent^{-1}[\{node\}] \land n\_tick(y) = TRUE \implies \\
                                n\_id(y) \leq n\_id(x)) \land n\_result(x) = SUCCESS)$\\
                    \hspace*{5pt}  $ \textbf{then} $\\
                    \hspace*{10pt} $ n\_result(node) := SUCCESS $ \\
                    \hspace*{10pt} $ analyzing\_subtree(parent(node)) := FALSE $ \\
                    \hspace*{5pt}  $ \textbf{end}\vspace{7pt} $ \\
                    
                    \hspace*{0pt}  $ \mathsf{fallback\_ticked\_continue}  \;\widehat{=}$ \vspace{1pt} \\
                    \hspace*{5pt} $\textbf{any} \; \; node, child$\\
                    \hspace*{5pt} $\textbf{where} $\\
                    \hspace*{10pt} $n\_type(node) = FALLBACK $\\
                    \hspace*{10pt} $n\_tick(node) = TRUE $\\
                    \hspace*{10pt} $n\_result(node) = UNKNOWN $\\
                    \hspace*{10pt} $\exists x \cdot (x \in  parent^{-1}[\{node\}] \land n\_tick(x) = TRUE \land \\
                                \forall y \cdot (y \in parent^{-1}[\{node\}] \land n\_tick(y) = TRUE \implies \\
                                n\_id(y) \leq n\_id(x)) \land n\_result(x) = FAILURE)$\\
                    \hspace*{10pt} $\{x | \; x \in  parent^{-1}[\{node\}] \; \\
                    \hspace*{10pt} \land \; n\_tick(x) = FALSE \} = \emptyset$ \\
                    \hspace*{10pt} $node \in nodes$\\
                    \hspace*{10pt} $child \in \{x \in  parent^{-1}[\{node\}] \\
                    \hspace*{10pt}\land n\_tick(x) = FALSE \land \\
                                \forall y \cdot (y \in parent^{-1}[\{node\}] \land n\_tick(y) = FALSE \implies \\
                                n\_id(y) \geq n\_id(x))$\\
                    \hspace*{10pt} $analyzing\_subtree(node) := FALSE$\\            
                    \hspace*{5pt} $\textbf{then}$\\
                    \hspace*{10pt} $n\_tick(child) := TRUE $ \\
                    \hspace*{10pt} $analyzing\_subtree(node) := TRUE $ \\
                    \hspace*{5pt} $\textbf{end}$\\
                    $\textbf{end}$
                }
            }
            
        }
        \end{small}
    \caption{The events to manage the Fallback node}
    \label{fig:fallback_events}
    \vspace*{-0.2cm}
    \end{center}
    \vspace*{-0.3cm}
\end{figure}

At every event either we return a result or we tick a child node. And we set the \textit{analyzing_subtree} label true or false accordingly.

For the \textbf{Sequence} node, we follow a similar approach we the modified event to control it according to its definition.\\
The definition is \textit{"The Sequence node [...], corresponds to routing the ticks to its children from the left until it finds a child that returns either Failure or Running, then it returns Failure or Running accordingly to its own parent. It returns success if and only if all its children return Success. Note that when a child returns Running or Failure, the Sequence node does not route the ticks to the next child (if any)."}~\cite{colledanchise_behavior_2018}. \\

The guards and events are similar to the Fallback node but with different logic. The events of the Sequence node can be seen in Figure~\ref{fig:sequence_events}. 

\begin{figure}[t!]
    \centering
    \begin{center}
        \begin{small}
        \fbox{
            \parbox[center]{15.7cm}{
                \hspace*{0pt} $\textbf{Events}\vspace*{7pt}$   \\
                \parbox[center]{8.25cm}{
                    \hspace*{0pt}  $ \mathsf{sequence\_ticked\_running}  \;\widehat{=}$ \vspace{1pt} \\
                    \hspace*{5pt}  $ \textbf{any} \; \; node$\\
                    \hspace*{5pt}  $ \textbf{where} $\\
                    \hspace*{10pt} $ n\_type(node) = SEQUENCE $\\
                    \hspace*{10pt} $ n\_tick(node) = TRUE $\\
                    \hspace*{10pt} $ n\_result(node) = UNKNOWN $\\
                    \hspace*{10pt} $ \exists x \cdot (x \in  parent^{-1}[\{node\}] \land n\_tick(x) = TRUE \land \\
                                \forall y \cdot (y \in parent^{-1}[\{node\}] \land n\_tick(y) = TRUE \implies \\
                                n\_id(y) \leq n\_id(x)) \land n\_result(x) = RUNNING)$\\
                    \hspace*{5pt}  $ \textbf{then} $\\
                    \hspace*{10pt} $ n\_result(node) := RUNNING $ \\
                    \hspace*{10pt} $ analyzing\_subtree(parent(node)) := FALSE $ \\
                    \hspace*{5pt}  $ \textbf{end}\vspace{7pt} $ \\
                    \hspace*{0pt}  $ \mathsf{sequence\_ticked\_failure}  \;\widehat{=}$ \vspace{1pt} \\
                    \hspace*{5pt} $\textbf{any} \; \; node$\\
                    \hspace*{5pt} $\textbf{where} $\\
                    \hspace*{10pt} $n\_type(node) = SEQUENCE $\\
                    \hspace*{10pt} $n\_tick(node) = TRUE $\\
                    \hspace*{10pt} $n\_result(node) = UNKNOWN $\\
                    \hspace*{10pt} $\exists x \cdot (x \in  parent^{-1}[\{node\}] \land n\_tick(x) = TRUE \land \\
                                \forall y \cdot (y \in parent^{-1}[\{node\}] \land n\_tick(y) = TRUE \implies \\
                                n\_id(y) \leq n\_id(x)) \land n\_result(x) = FAILURE)$\\
                    \hspace*{5pt} $\textbf{then}$\\
                    \hspace*{10pt} $n\_result(node) := FAILURE$ \\
                    \hspace*{10pt} $analyzing\_subtree(parent(node)) := FALSE$ \\
                    \hspace*{5pt} $\textbf{end}$\\
                }
                \parbox[center]{8cm}{
                    \hspace*{0pt}  $ \mathsf{sequence\_ticked\_success}  \;\widehat{=}$ \vspace{1pt} \\
                    \hspace*{5pt}  $ \textbf{any} \; \; node$\\
                    \hspace*{5pt}  $ \textbf{where} $\\
                    \hspace*{10pt} $ n\_type(node) = SEQUENCE $\\
                    \hspace*{10pt} $ n\_tick(node) = TRUE $\\
                    \hspace*{10pt} $ n\_result(node) = UNKNOWN $\\
                    \hspace*{10pt} $ \exists x \cdot (x \in  parent^{-1}[\{node\}] \land n\_tick(x) = TRUE \land \\
                                \forall y \cdot (y \in parent^{-1}[\{node\}] \land n\_tick(y) = TRUE \implies \\
                                n\_id(y) \leq n\_id(x)) \land n\_result(x) = SUCCESS)$\\
                    \hspace*{10pt} $\{x | \; x \in  parent^{-1}[\{node\}] \; \\
                    \hspace*{10pt}\land \; n\_tick(x) = FALSE \} = \emptyset$ \\
                    \hspace*{10pt} $node \in nodes$\\
                    \hspace*{5pt}  $ \textbf{then} $\\
                    \hspace*{10pt} $ n\_result(node) := SUCCESS $ \\
                    \hspace*{10pt} $ analyzing\_subtree(parent(node)) := FALSE $ \\
                    \hspace*{5pt}  $ \textbf{end}\vspace{7pt} $ \\
                    
                    \hspace*{0pt}  $ \mathsf{sequence\_ticked\_continue}  \;\widehat{=}$ \vspace{1pt} \\
                    \hspace*{5pt} $\textbf{any} \; \; node,child$\\
                    \hspace*{5pt} $\textbf{where} $\\
                    \hspace*{10pt} $n\_type(node) = SEQUENCE $\\
                    \hspace*{10pt} $n\_tick(node) = TRUE $\\
                    \hspace*{10pt} $n\_result(node) = UNKNOWN $\\
                    \hspace*{10pt} $\exists x \cdot (x \in  parent^{-1}[\{node\}] \land n\_tick(x) = TRUE \land \\
                                \forall y \cdot (y \in parent^{-1}[\{node\}] \land n\_tick(y) = TRUE \implies \\
                                n\_id(y) \leq n\_id(x)) \land n\_result(x) = SUCCESS)$\\
                    \hspace*{10pt} $\{x | \; x \in  parent^{-1}[\{node\}] \; \\
                    \hspace*{10pt}\land \; n\_tick(x) = FALSE \} = \emptyset$ \\
                    \hspace*{10pt} $node \in nodes$\\
                    \hspace*{10pt} $child \in \{x \in  parent^{-1}[\{node\}] \\
                    \hspace*{10pt}\land n\_tick(x) = FALSE \land \\
                                \forall y \cdot (y \in parent^{-1}[\{node\}] \land n\_tick(y) = FALSE \implies \\
                                n\_id(y) \geq n\_id(x))$\\
                    \hspace*{10pt} $analyzing\_subtree(node) := FALSE$\\            
                    \hspace*{5pt} $\textbf{then}$\\
                    \hspace*{10pt} $n\_tick(child) := TRUE $ \\
                    \hspace*{10pt} $analyzing\_subtree(node) := TRUE $ \\
                    \hspace*{5pt} $\textbf{end}$\\
                    $\textbf{end}$
                }
            }
            
        }
        \end{small}
    \caption{The events to manage the Sequence node}
    \label{fig:sequence_events}
    \vspace*{-0.2cm}
    \end{center}
    \vspace*{-0.3cm}
\end{figure}

\subsection{Instantiating generic specification}
The fourth refinement finally allows us to deal with real instances of BTs. At this level, actual Behaviour Trees can be verified and checked for invariants of interest for the agent. We now explain how the context and the machine need to be built to properly check an existing tree.

The context contains the static part of the machine, the BT in itself is a static part of the agent, and for this reason, the definition of the topology and of the nodes type of the BT needs to be defined here.

To define a BT instance we start by extending the \textit{BT_level} and listing all the nodes contained in the tree (excluding the root, already defined in the previous refinement), as constants.

We then need to define the axioms regarding the type: that every node defined in the constant part is of type \textit{nodes}. 
After this we need to initialize all the nodes giving to them a \textit{n_type}, and a \textit{n_id}. Finally, we construct the parent function, defining for each node which node is its parent, defining in this way the topology of the tree.

Numbering the nodes correctly is essential to correctly navigate a tree.
To correctly give a \textit{n_id} to the nodes, is required to follow the same principle as a breadth-first search: firstly considering the depth, and then starting counting the children from left to right. Therefore, The root, receive the number $0$. We will then give to all its children the next numbers starting by counting from the left-most node (i.e. the first to be ticked). When all the children have been assigned a number we start to analyze the children of the first child, numbering them left-to-right in the same way as before. And so on, until all the nodes have been numbered.


Moving on the machine of the BT instance, we start as a refinement of the \textit{BT_machine} of the previous level.
The only effort required here is the extension and refinement of the events of the leaf nodes. Since the leaf nodes (conditions and actions) are very specific to the application, we need to extend the corresponding events for any action and condition we want to check.
For instance, let's suppose that our agent has two condition nodes: \textit{condition_1} and \textit{condition_2}, the first condition checks if \emph{the agent is more than 3 meters from an obstacle}, and the second that \emph{the battery level is more than 30\%}. 
To cover the two conditions we would need to extend the \textit{condition_ticked} event of the BT level four times:
\begin{enumerate*}
 \item \textit{condition_1_ticked_success}, 
 \item \textit{condition_1_ticked_failure}, 
 \item \textit{condition_2_ticked_success}, 
 \item \textit{condition_2_ticked_failure}, 
\end{enumerate*}
covering all the possible results that the nodes can output. The modifications to be added would be minimal, in the \textit{condition_1} case we would just need a guard to check the condition (something like $distance\_to\_obstacle \geq 3$) and an action that explain which value to return ($ n\_result(node) := SUCCESS $). For the failure case, we would add the condition for failure ($distance\_to\_obstacle < 3$) and the corresponding returning value ($ n\_result(node) := FAILURE $)

Moreover, if we want to check some safety invariant or other parameters, we would need to model the part of the environment we are interested to check and how the various actions would change the environment itself. For instance, if a robot is moving in an environment with some obstacles we could define two variables (\textit{pos_x}, \textit{pos_y} to define the position of the robot and another variable (\textit{distance_to_obstacle}), to keep track of the distance to the closest obstacle. These would then be modified by actions. For instance, if we have an action "move one step up", in the event of the corresponding action we would also need to add something like: $pos\_y := pos\_y + 1$.

\section{Case Study -- A moving robot}

\begin{figure}[t!]
    \centering
    \begin{center}
    \includegraphics[width=14.8cm]{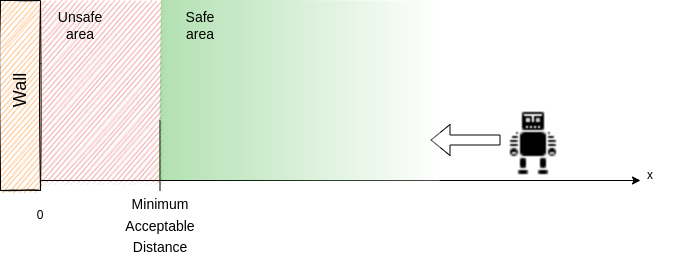}
    \caption{Case study of a robot moving towards a wall, without going too close to it. The robot can move closer to the wall, but it is not allowed to enter the unsafe area.}
    \label{fig:robot_to_wall}
    \vspace*{-0.2cm}
    \end{center}
    \vspace*{-0.3cm}
\end{figure}

As a case study, we analyzed a very simple case, considering a robot that is moving toward a wall. The goal is to avoid the robot from going too close to a certain distance to the obstacle Fig: \ref{fig:robot_to_wall}.

To model this simple environment we have defined in the machine different variables: 
\begin{enumerate*}
    \item \textit{distance_to_object}, 
    \item \textit{time}, 
    \item \textit{prev_time}
\end{enumerate*}. 
We use the \textit{distance_to_object} variable to identify the position of the robot and hence its distance to the wall, while a \emph{time} variable (together with the previous time step) is used to control the real-time passed during the execution.

\begin{figure}
    \centering
    \includegraphics[width=9cm]{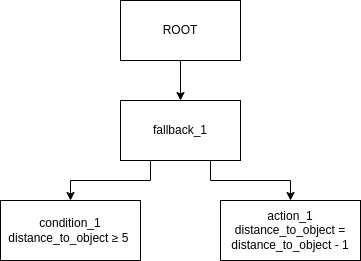}
    \caption{Behaviour tree used to model the decision-making process of the agent.}
    \label{fig:case_study_BT}
\end{figure}

To define the behavior of the robot we have defined BT represented in Figure~\ref{fig:case_study_BT}.

As invariant to be checked we set:
\begin{equation*}
    distance\_to\_object \geq 3
\end{equation*}

In the initialization, we set the robot at a distance of 10 meters from the wall.

The first leaf node of the tree is a condition that the robot needs to be at least 5 meters from the wall. If this condition holds then the robot can perform the action of moving one step closer to the wall.

To model this environment, we followed the same approach described at the instance level above. 
We first defined the tree in the context described below. In the machine, we define the variables described above to define the environment and we refine the leaf node events.

For the condition, we create the two refinements of the event for the $SUCCESS$ and $FAILURE$ case as described in the methodology section. For the action, we simplify the scenario enforcing that, once selected, the action will take effect instantly and that can never fail. Allowing us to refine the action event just for the $SUCCESS$ case. 

\begin{figure}[H]
    \centering

\begin{center}
\begin{small}
\fbox{
	\parbox[center]{15.2cm}{
    \vspace{1pt}
    \hspace*{0pt} $\textbf{context}\; robot\_wall\_context \;$\\
    \hspace*{0pt} $\textbf{extends}\; 03\_BT\_context\;$\\

	\hspace*{0pt} $\textbf{constants}\; \;$\\
	\hspace*{5pt} $sequence\_1$\\
	\hspace*{5pt} $condition\_1$\\
	\hspace*{5pt} $action\_1$\\
	\hspace*{0pt} $\textbf{axioms} \; \;$\\
	\hspace*{5pt} $sequence\_1 \in nodes $\\
	\hspace*{5pt} $condition\_1 \in nodes $\\
	\hspace*{5pt} $action\_1 \in nodes $\\
    \hspace*{5pt} $n\_type = \{root \mapsto ROOT, \; sequence\_1 \mapsto SEQUENCE, \; condition\_1 \mapsto CONDITION, \; action\_1 \mapsto ACTION\}  $\\
    \hspace*{5pt} $n\_id = \{root \mapsto 0, \; sequence\_1 \mapsto 1, \; condition\_1 \mapsto 2, \; action\_1 \mapsto 3\}$ \\
	\hspace*{5pt} $parent = \{sequence\_1 \mapsto root, \; condition\_1 \mapsto sequence\_1, \; action\_1 \mapsto sequence\_1\}  $\\
	}
 }
	
\end{small}
\end{center}
\caption{The context of the BT for the robot moving towards the wall}
\end{figure}

We can note that the action ticked event is very similar to the previous definition, with the only difference due to the addition of the last two propositions.

\begin{figure}[H]
    \centering
\begin{center}
\begin{small}
\fbox{
	\parbox[center]{15.2cm}{
    \vspace{1pt}
    \hspace*{0pt}  $ \mathsf{action\_ticked\_success}  \;\widehat{=}$ \vspace{1pt} \\
    \hspace*{5pt}  $ \textbf{REFINES} \; action\_ticked$\\
    \hspace*{5pt}  $ \textbf{any} \; \; node$\\
    \hspace*{5pt}  $ \textbf{where} $\\
    \hspace*{10pt} $ n\_type(node) = ACTION \; \text{// not changed}$\\
    \hspace*{10pt} $ n\_tick(node) = TRUE \; \text{// not changed} $\\
    \hspace*{10pt} $ n\_result(node) = UNKNOWN \; \text{// not changed} $\\
    \hspace*{10pt} $ node \in nodes \; \text{// not changed} $\\
    \hspace*{5pt}  $ \textbf{then} $\\
        \hspace*{10pt} $ analyzing\_subtree(parent(node)) := FALSE \; \text{// not changed} $ \\
        \hspace*{10pt} $ n\_result(node) := SUCCESS \; \text{// added in refinement} $ \\
        \hspace*{10pt} $ distance\_to\_object := distance\_to\_object -1 \; \text{// added in refinement} $ \\
    \hspace*{5pt}  $ \textbf{end}\vspace{7pt} $
	}}
	
\end{small}
\end{center}
\caption{The context of the BT for the robot moving towards the wall}
\label{evt:action_ticked}
\end{figure}
We then added a statement to increase the timestep whenever the result reaches the root node, considering an entire exploration of the tree as a timestep in the real world since the changes in the position of the robot have been applied to the action event.

In general, just with these modifications at the machine from the BT level, we are able to ensure the safety properties of the system by proving the safety invariant that we have defined for this BT. This ensures us that, after the initialization, the safety statement that we have described will hold for each possible state that the BT can reach. This can be expanded, with multiple safety statements and multiple requirements, allowing us to formally prove the safety of the system that will implement this specific BT.

\section{Related Works}

Although Behavior Trees have started to become widely used in various robotics applications, still little research has been done to increase the dependability of robots using them. The importance of providing safe and reliable agents becomes more and more critical, especially given the fact that these tools are starting to be used at a larger level, especially with the support of open source software like the Nav2 package for ROS~\cite{macenski_marathon_2020}. 
Work in the safety assessment of Behavior Trees has been done, for instance, \cite{lindsay_safety_2010}, but these studies normally refer to specific cases and become difficult to generalize for other BT. BT can also be safely generated as in~\cite{9030183,10.1109/IROS.2017.8206502} and that BT is an optimal tool to build safe and reliable systems \cite{6942752}. However, formal verification of safety properties is rarely done on new trees especially in manually constructed ones, risking in this way the overall dependability of the systems using them.
Other works propose the formal verification of BT using Linear Temporal Logic \cite{biggar_framework_2020},  but they do not provide a tool to support the verification process. 

Event-B was used for modelling robotic systems by Troubitsyna et al. \cite{LAIBINIS201766, VisTro21, PTL2012}. These works adopted goal-oriented model as a basis of robot’s decision making and demonstrated goal reachability despite robot’s failures. However, these works rely on an assumption that the robot’s operating environment remains static, i.e., once a certain goal is accomplished, its status remains unchanged throughout the entire system execution. In our work, reliance on BT allows us to adopt dynamic model of system environment, because at each tick the status of all tasks is getting checked, i.e., this work offers a more flexible basis for modeling robot’s decision making. 

\section{Conclusion}
In this work, we present a formal specification of Behaviour Trees in Event-B. We also provide a methodology to test new instances of BT in a fairly simple manner for newcomers hiding the definition of the BT at more abstract levels, allowing a user to focus just on test and checking the properties the specific BT needs to hold. Our approach aimed to facilitate the formal verification of Behaviour Trees without the need to build the formalization of the tree and the control flow node from scratch and simplifying also the modeling of the leaf nodes.

In future works, we propose to improve the actual specification by adding the support for parallel nodes and the most common decorator used in practice. 
Moreover, we would like to simplify further the possibility to check already used BT, creating a tool to automatically import trees built using already used libraries like py\_tree \cite{splintered-reality} or Behaviour\_Tree.CPP \cite{behaviortree}. Of particular interest is also the analysis of already used BT in robotic frameworks like the ones used in the Navigation 2 \cite{macenski_marathon_2020} package in ROS.

\nocite{*}
\bibliographystyle{eptcs}
\bibliography{generic}

\end{document}